\pdfoutput=1

\documentclass[11pt]{article}

\usepackage[final]{acl}

\usepackage{times}
\usepackage{latexsym}

\usepackage{booktabs,multirow,longtable, float}
\usepackage{amsmath}

\usepackage[T1]{fontenc}

\usepackage[utf8]{inputenc}

\usepackage{microtype}

\usepackage{inconsolata}

\usepackage{graphicx}

\title{The Fellowship of the LLMs: Multi-Model Workflows for Synthetic Preference Optimization Dataset Generation}

\author{
Samee Arif\textsuperscript{\rm 1}\textsuperscript{*}, Sualeha Farid\textsuperscript{\rm 2}\textsuperscript{*}, Abdul Hameed Azeemi\textsuperscript{\rm 1} \\ \textbf{Awais Athar}\textsuperscript{\rm 3,4}\textsuperscript{\textdagger}, \textbf{Agha Ali Raza}\textsuperscript{\rm 1} \\
\textsuperscript{\rm 1}Lahore University of Management Sciences,
\textsuperscript{\rm 2}University of Michigan - Ann Arbor \\
\textsuperscript{\rm 3}EMBL European Bioinformatics Institute,
\textsuperscript{\rm 4}Strategize Inc \\
\texttt{\{samee.arif, abdul.azeemi, agha.ali.raza\}@lums.edu.pk}\\
\texttt{sualeha@umich.edu}, \texttt{awais@strategize.inc}
}

\begin{document}
\maketitle
\renewcommand{\thefootnote}{\fnsymbol{footnote}}
\footnotetext[1]{These authors contributed equally to this work.}
\renewcommand{\thefootnote}{\textdagger}
\footnotetext[1]{Work done while at EMBL-EBI}
\renewcommand{\thefootnote}{\arabic{footnote}}
\maketitle
\begin{abstract}
This paper presents a novel methodology for generating synthetic Preference Optimization (PO) datasets using multi-model workflows. We evaluate the effectiveness and potential of these workflows in automating and enhancing the dataset generation process. PO dataset generation requires two modules: (1) \textit{response evaluation}, and (2) \textit{response generation}. In the \textit{response evaluation} module, the responses from Large Language Models (LLMs) are evaluated and ranked - a task typically carried out by human annotators that we automate using LLMs. We assess the response evaluation module in a 2 step process. In step 1, we assess LLMs as evaluators using three distinct prompting strategies. In step 2, we apply the winning prompting strategy to compare the performance of LLM-as-a-Judge, LLMs-as-a-Jury, and LLM Debate. Our evaluation shows that GPT-4o-as-a-Judge is more consistent across all datasets. For the \textit{response generation} module, we use the identified LLM evaluator configuration and compare different configurations of the LLM Feedback Loop. We use the win rate to determine the best multi-model configuration for generation. Experimenting with various configurations, we find that the LLM Feedback Loop, with Llama as the generator and Gemma as the reviewer, achieves a notable 71.8\% and 73.8\% win rate over single-model Llama and Gemma, respectively. After identifying the best configurations for both modules, we generate our PO datasets using the above pipeline.
\end{abstract}

\section{Introduction}

Large Language Models (LLMs) demonstrate a range of Natural Language Processing (NLP) capabilities, including text generation, question answering, and language understanding. However, LLMs can sometimes deviate from user instructions and exhibit unintended behaviors \cite{tamkin2021understanding}. To mitigate this problem and align the LLM outputs more closely with human preferences, techniques like Reinforcement Learning from Human Feedback (RLHF) are used, which involves fine-tuning LLMs using the reward signal from human preferences \cite{christiano2017deep}. Improved methods like Direct Preference Optimization (DPO) \cite{rafailov2024directpreferenceoptimizationlanguage} eliminate the need for fitting the reward model and are more stable and performant. In DPO, the preference optimization dataset requires a pair of accepted and rejected responses for each prompt. The accepted response is one that better aligns with the desired human preferences. Other techniques like Kahneman-Tversky Optimization (KTO) \citep{ethayarajh2024ktomodelalignmentprospect} require each response to indicate whether it is good or bad (i.e., as a binary classification task) instead of pairwise preferences.

In the process of constructing the dataset of human preferences, the evaluation and ranking of the outputs generated by LLMs are typically done by human annotators, who assess these outputs based on various criteria such as instruction following, helpfulness, relevance, accuracy, depth, and creativity. The PO dataset generation process is divided into two modules: response evaluation and response generation. The response evaluation module involves assessing and ranking responses generated by LLMs, while the response generation module focuses on creating responses that align with the identified preferences. This manual process, while effective, is labor-intensive, time-consuming, inconsistent, and subject to human biases. In this work, we thus ask the question, \textit{Can we use LLMs to automate and improve response evaluation and generation for constructing preference optimization (PO) datasets?}. 

For the response evaluation step, we leverage LLMs as evaluators and compare several configurations including LLM-as-a-Judge, LLMs-as-a-Jury, and LLM Debate to pick the best evaluation strategy. The selected response evaluation module is used to evaluate and identify the optimal response generation module. Previously, single-models have been used to generate the responses for PO datasets; however, we use a multi-model framework for response generation, which allows us to generate more refined, higher-quality responses. The multi-model approach uses the collaboration between multiple LLMs, where one model can provide suggestions for improvements, and the other can revise the response based on the feedback. This iterative process leads to a thorough refinement of the generated content, ensuring that the final output better aligns with human preferences and expectations.

In this framework, the response generation module produces several possible responses, and the response evaluation module selects the best one from the list to create the PO dataset. We present multiple DPO and KTO datasets with the focus is on generating datasets to improve the performance of individual LLMs. The primary aim of the datasets is to enhance the performance and capabilities of individual LLMs by providing high-quality PO training data that better aligns with human judgment and expectations. Our contributions can be summarized as follows:


\section{Related Work}

\subsection{Preference Optimization}
Preference Optimization has emerged as a pivotal technique for aligning model outputs with human preferences. \citet{rafailov2024directpreferenceoptimizationlanguage} introduce DPO, a method that simplifies solving the standard RLHF problem by converting it into a classification task, enabling the extraction of the optimal policy in a straightforward way. \citet{hong2024orpomonolithicpreferenceoptimization} introduce ORPO algorithm that combines the traditional supervised fine-tuning and preference alignment stages into a single process. The dataset for DPO and ORPO require annotated preference pairs, where each pair consists of two model outputs labeled according to which one better aligns with human preferences. \citet{ethayarajh2024ktomodelalignmentprospect} introduce KTO, a cost-effective approach to align Large Language Models (LLMs) with human feedback, improving performance without the need for preference pairs. Argilla Distilabel \citep{distilabel-argilla-2024} uses LLM to judge between the responses of two models to create synthetic PO datasets. The datasets are available on Hugging Face\footnote{\url{https://huggingface.co/argilla}}. To our knowledge, no one has yet explored the use of multi-model workflows for the generation of PO datasets.

\subsection{Multi-Model Frameworks}
Recently, there has been a growing interest in using LLM multi-model frameworks for different tasks. \citet{zheng2023judgingllmasajudgemtbenchchatbot} presents an evaluation of LLM-as-a-Judge on the MT-Bench \citep{zheng2023judging} and Chatbot Arena \citep{li2024crowdsourceddatahighqualitybenchmarks}. Their results reveal that strong LLM judges like GPT-4 can match both controlled and crowd-sourced human preferences well, achieving over 80\% agreement, the same level of agreement between humans. Additionally, they evaluate several variants of Llama and Vicuna on the dataset. They study the limitations of LLM-as-a-judge, including position, verbosity, and self-enhancement biases, as well as limited reasoning ability. \citet{verga2024replacingjudgesjuriesevaluating} explore the use of LLMs-as-a-Jury. Their approach, a Panel of LLM evaluators (PoLL), composed of a larger number of smaller models outperforms a single large judge. They also show that the PoLL approach exhibits less intra-model bias as compared to LLM-as-a-Judge. They use Command-R, GPT, Claude-3, and Mistral families for their study. Additionally, they compare two prompting strategies: (1) reference-based scoring where they provide the LLM with a reference answer, and (2) candidate answer and pair-wise scoring where they ask the LLM to pick the better response from the candidate responses. PoLL outperforms single-models on KILT \citep{petroni2021kiltbenchmarkknowledgeintensive} and Chatbot Arena.

\citet{liang2024encouragingdivergentthinkinglarge} introduce Multi-Agent Debate (MAD) to encourage divergent thinking in LLMs. They mitigate the Degeneration-of-Thought (DoT) problem, which is that once the LLM has established confidence in its solutions, it is unable to generate novel thoughts. In their approach, the affirmative LLM and the negative LLM debate on the answer while the LLM judge evaluates both arguments after each round of debate. They evaluate the approach on the Commonsense Machine Translation Dataset (Chinese to English) \citep{he-etal-2020-box} and their Counter-Intuitive Arithmetic Reasoning (CIAR) dataset. MAD was able to achieve a 37\% accuracy on the CIAR dataset using GPT-3.5-Turbo which outperforms Chain-of-Thought, Self-Consistency, and Self-Reflection prompting. They also show that using the MAD approach decreases bias and increases response diversity. \citet{du2023improvingfactualityreasoninglanguage} evaluates a different variant of multi-model debate where multiple models generate their own responses, and each model receives the opinions of the other models, then updates its response if necessary. This is done for multiple rounds. \citet{du2023improvingfactualityreasoninglanguage} evaluates the approach on the following tasks: Biography generation, MMLU, Chess move validity and optimality, Arithmetic, and Grade school math,. Their approach using ChatGPT and Bard outperforms single-model on all the tasks. To evaluate LLM responses \citet{chan2023chatevalbetterllmbasedevaluators} presents another variant of multi-model debate. Their architecture involves assigning models different roles such as General Public, Critic, Psychologist, News Author, and Scientist. They used ChatGPT and GPT-4 for their evaluation on FairEval \citep{Wang2023LargeLM} dataset and achieved a Cohen's Kappa score of 0.40 using LLM Debate, 0.03 more than the single-model.

\section{Methodology}

\subsection{Experimental Setup}
In this study, we perform experiments on the three categories of LLMs given in Table \ref{tab:models}. For the evaluation module, we evaluate single-models and multi-model frameworks on four datasets, Alpaca Eval \citep{alpaca_eval}, FairEval \citep{Wang2023LargeLM}, PandaLM-Eval \citep{pandalm2024,PandaLM} and MT-Bench \citep{zheng2023judging}. For the generation module, we compare the multi-model frameworks using win rate - the ratio of times a generation framework is selected as the best by an LLM evaluator when comparing outputs from all generation workflows. After the extensive evaluation of both modules, we used the picked strategies to generate synthetic PO datasets. We set the temperature to 0 in all our evaluations to ensure reproducibility.

\begin{table}[h]
\centering
\small
\begin{tabular}{ll}
\toprule
\textbf{Category} & \textbf{Models} \\ \midrule
\multirow{2}{*}{Small-Scale LLM}
  & Llama-3.1-8b \\
  & Gemma-2-9b \\[.5em]
\multirow{2}{*}{Mid-Scale LLM}
  & Gemma-2-27b \\
  & Llama-3.1-70b \\[.5em]
\multirow{2}{*}{Large-Scale LLM}
  & GPT-4o-Mini (2024-07-18) \\
  & GPT-4o (2024-05-13) \\
\bottomrule
\end{tabular}
\caption{\small{Categories of LLMs used in the study.}}
\label{tab:models}
\vskip -0.1in
\end{table}

\subsection{LLM-as-Evaluator}
With the aim of automating the evaluation component of PO dataset generation, we assess the performance of LLMs in the role of evaluators using the Alpaca Eval, FairEval, PandaLM-Eval, and MT-Bench datasets. Our goal is to determine whether multi-model workflows work better than a single-model for LLM evaluation. The system prompts for this task are modified version of the prompts used by \citet{zheng2023judgingllmasajudgemtbenchchatbot} and are given in Appendix \ref{sec:system-prompts}.

\paragraph{LLM-as-Judge.}
We evaluate six different LLMs on the Alpaca Eval dataset, calculating Cohen’s Kappa with the human annotations. Our evaluation involved three distinct prompting strategies for the LLM-as-a-Judge:
\begin{enumerate}
    \item \textbf{Direct Comparison:} The Judge-LLM is provided with the user question and the responses generated by different LLMs. It is asked to pick the best response among the given options.
    \item \textbf{Independent Scoring:} The Judge-LLM is given the user question and each response in separate conversations. It is asked to score each response independently.
    \item \textbf{Combined Scoring:} The Judge-LLM is provided with the user question and all the responses in a single conversation thread. It is asked to assign a score to each response within the same conversation context. To observe if the scoring range influences the LLM’s scoring consistency and its alignment with human annotations, we test three different scoring totals: 5, 10, and 100.
\end{enumerate}
For each of these prompting strategy, we systematically analyze the performance of the LLMs by calculating Cohen’s Kappa, against the human annotations. The system prompts are given in Table \ref{tab:prompt-llm-judge} in Appendix \ref{sec:system-prompts}.

\paragraph{LLMs-as-Jury.}
We extend the evaluation from the LLM-as-a-Judge approach by forming juries composed of multiple LLMs. Specifically, we test all possible combinations of the six LLM models when forming juries of sizes ranging from 2 to 6. We use three datasets: FairEval, PandaLM-Eval and MT-Bench datasets for a more comprehensive analysis. We systematically analyze the performance of each jury configuration, focusing on how the size and combination of the LLMs affect their judgment accuracy. The \textit{Combined Scoring} system prompt in Table \ref{tab:prompt-llm-judge} in Appendix \ref{sec:system-prompts} is used for all the jurors because it performed the best in our previous evaluation.

\begin{figure}[h!]
    \begin{center}
    \centerline{\includegraphics[width=0.9\columnwidth]{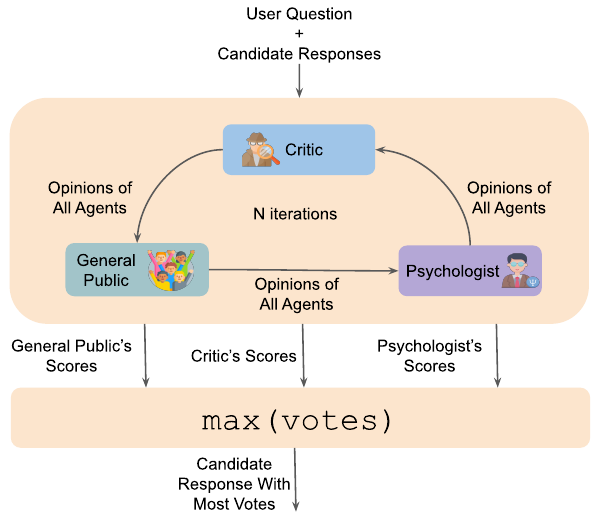}}
    \caption{\small{LLM Debate for evaluation}}
    \label{fig:evaluation-debate}
    \end{center}
    \vskip -0.3in
\end{figure}

\paragraph{LLM Debate.}
We also evaluate the LLM Debate framework following the implementation described by \citet{chan2023chatevalbetterllmbasedevaluators}. In this approach, we assign three distinct roles—Psychologist, General Public, and Critic—and the three models debate the scores that should be assigned to candidate responses. After the debate, each model gives its final score which is used to determine which candidate response they vote for. These votes are then used to pick the best response. This strategy is evaluated using the FairEval, PandaLM-Eval, and MT-Bench benchmarks. Figure \ref{fig:evaluation-debate} illustrates the debate workflow employed in our study. The system prompt, the user message structure and the prompts for the roles used are given in Table \ref{tab:prompt-llm-debate} and Table \ref{tab:prompt-roles} in Appendix \ref{sec:system-prompts}.

\subsection{LLM-as-Generator}
\begin{figure}[h!]
    \begin{center}
    \centerline{\includegraphics[width=0.8\columnwidth]{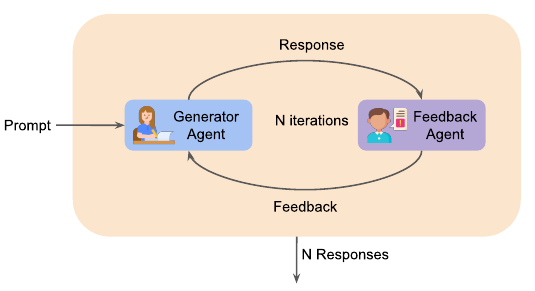}}
    \caption{\small{LLM Feedback Loop for response generation}}
    \label{fig:generation-feedback}
    \end{center}
    \vskip -0.3in
\end{figure}

To evaluate the LLM Feedback Loop workflow for the generation module, we test different configurations using Llama-3.1-8b \citep{meta2024llama3} and Gemma-2-9b \citep{google2024gemma2} models. In this framework, a generator LLM produces a response, which is then evaluated by a feedback LLM that provides improvement suggestions as shown in Figure \ref{fig:generation-feedback}. The generator revises the response based on these suggestions, and the process repeats for multiple iterations. The system prompt for the generator and reviewer is given in Table \ref{tab:generator_prompt} and \ref{tab:reviewer_prompt} in Appendix \ref{sec:system-prompts}. We calculate the win rate against single-model GPT-4o \citep{openai2024gpt4}, Llama-3.1-8b and Gemma-2-9b baseline outputs on a subset of 500 prompts from the Argilla Capybara DPO dataset\footnote{\url{https://huggingface.co/datasets/argilla/distilabel-capybara-dpo-7k-binarized}} to identify the best configuration. We test the following configuration:
\begin{enumerate}
    \item \textbf{Same Model:} Gemma-2-9b or Llama-3.1-8b as both the feedback and generation model.
    \item \textbf{Different Models:} Gemma-2-9b as the feedback model and Llama-3.1-8b as the generation model, or vice versa.
    \item \textbf{Both Models for Feedback, One for Generation:} Gemma-2-9b or Llama-3.1-8b as the generation model, with both models as feedback model.
\end{enumerate}

\subsection{Preference Optimization Dataset}
We use the best configurations of the generation and evaluation modules to generate the DPO and KTO datasets. The generation module produces $N$ responses (where $N$ is the number of feedback iterations), which are then passed to the evaluation module. The evaluation module sorts these responses into the accepted and rejected fields in the DPO and KTO datasets. In this study, we use the prompts from the Argilla Capybara DPO dataset. The prompt templates used for LLM improvement dataset generation are given in Table \ref{tab:prompt-llm-judge}, \ref{tab:generator_prompt} and \ref{tab:reviewer_prompt} in Appendix \ref{sec:system-prompts}. The evaluation code, all the evaluation outputs and the generated datasets are publicly available on GitHub\footnote{\url{https://github.com/sameearif/Fellowship-of-The-LLMs}}.

\section{Results and Discussion}
\subsection{LLM-as-Evaluator}

\paragraph{Prompting Strategies.} Table \ref{tab:promptresults} shows the results of LLM-as-a-Judge approach on the three prompting strategies.

\begin{center}
\begin{table}[h!]
\centering
\renewcommand{\arraystretch}{1}
\setlength{\tabcolsep}{4.7pt}
\renewcommand{\arraystretch}{1}
\small
\begin{tabular}{lcccccccccc}
\toprule
& \textbf{Comp.} & \multicolumn{1}{c}{\textbf{Ind.}} & \multicolumn{3}{c}{\textbf{Combined}} \\
\cmidrule(lr){4-6}
\textbf{Judge} & & \textbf{10} & \textbf{5} & \textbf{10} & \textbf{100}  \\
\midrule
Gemma-2-9b & 0.226 & 0.170 & 0.243 & 0.254 & 0.233 \\
Llama-3.1-8b & 0.265 & 0.181 & 0.255 & 0.240 & 0.242 \\
Gemma-2-27b & 0.233 & 0.173 & 0.284 & 0.266 & 0.252 \\
Llama-3.1-70b & 0.305 & 0.214 & 0.337 & 0.333 & 0.339 \\
GPT-4o-mini & 0.342 & \textbf{0.254} & 0.374 & \textbf{0.382} & 0.347 \\
GPT-4o & \textbf{0.372} & 0.249 & \textbf{0.393} & \textbf{0.382} & \textbf{0.401} \\
\bottomrule
\end{tabular}
\caption{\small{Performance comparison of LLM-as-a-Judge on Alpaca-Eval using different prompting strategies. Direct Comparison (\textbf{Comp.}) vs. Independent Scoring (\textbf{Ind.}) vs. Combined Scoring (\textbf{Combined}). The bold values indicate the highest Cohen's kappa values for a particular strategy.}}
\label{tab:promptresults}
\end{table}
\vskip -0.2in
\end{center}

The \textit{Independent Scoring} prompt strategy consistently under-performs compared to the \textit{Direct Comparison} and \textit{Combined Scoring} approaches across all evaluated LLMs. This result is reflected in lower Cohen’s Kappa values ranging from only 0.170 to 0.254 in Table \ref{tab:promptresults}. In evaluating responses in isolation the LLM has to re-calibrate its scoring mechanism for every new response. This can lead to inconsistencies, especially when multiple responses are closely matched in quality. Due to the low Kappa values observed, we opted not to conduct experiments with the scoring-out-of-5 and 100 scales for \textit{Independent Scoring}.

The \textit{Direct Comparison} Strategy performs better than the \textit{Independent Scoring} approach across most LLMs, with a notable improvement for GPT-4o (0.372 vs. 0.249) and GPT-4o-mini (0.342 vs. 0.254). However, it generally falls short when compared to the \textit{Combined Scoring} method, where GPT-4o achieves a score of 0.401 using the scoring-out-of-100 scale. The higher Cohen’s Kappa values indicate that the \textit{Direct Comparison} and \textit{Combined Scoring} strategy benefits from providing the LLM with a side-by-side evaluation of responses, allowing for more accurate and consistent judgments.

The \textit{Combined Scoring} strategy, as presented in Table \ref{tab:promptresults}, shows consistent performance using all the scoring scales. It outperforms both the other prompts. The scoring scales of 5, 10, and 100 show variability across different models, with certain scales performing better for some models than others. For example, GPT-4o performs the best in scoring-out-of-10 scale with a Kappa score of 0.382 while Gemma-2-9b performs best under scoring-out-of-5 scale. Given these results, we selected the scoring-out-of-10 scale as the most effective option for the \textit{Combined Scoring} approach. We use this prompt for all our further evaluations.

\paragraph{LLM-as-a-Judge.} The LLM-as-Judge evaluations, as shown in Table \ref{tab:promptresults}, indicate that GPT-4o outperforms all the models on PandaLM-Eval and MT-Bench achieving a Cohen's Kappa score of 0.688 and 0.410 respectively. Additionally, GPT-4o consistently ranks in second position across all three datasets. This consistent top-tier performance underscores GPT’s effectiveness as a reliable judge in evaluating LLM responses. Gemma-2-27b outperforms all other models on the Fair-Eval dataset, achieving the highest score in this particular evaluation. However, it’s important to note that the Fair-Eval dataset is relatively small, consisting of only 80 samples. Furthermore, the Fair-Eval dataset primarily compares GPT-3.5-Turbo with Vicuna-13b, which might introduce a bias in favor of GPT models when GPT is the evaluator. 

\begin{center}
\begin{table}[htbp]
\centering
\renewcommand{\arraystretch}{1}
\small 
\begin{tabular}{lcccccccccc}
\toprule
& \textbf{Fair-Eval} & \textbf{PandaLM} & \textbf{MT-Bench} \\
\textbf{Judge}  \\
\midrule
Gemma-2-9b & 0.279 & 0.595 & 0.354 \\
Llama-3.1-8b & 0.206 & 0.523 & 0.339 \\
Gemma-2-27b & \textbf{0.389} & 0.586 & 0.354 \\
Llama-3.1-70b & 0.257 & 0.597 & 0.387 \\
GPT-4o-mini & 0.333 & 0.613 & 0.388 \\
GPT-4o & 0.327 & \textbf{0.688} & \textbf{0.410} \\
\bottomrule
\end{tabular}
\caption{\small{Performance comparison of LLM-as-a-Judge on Alpaca-Eval using different prompting strategies. Direct Comparison vs. Independent Scoring (out of 10) vs. Combined Scoring (out of 5, 10 and 100).}}
\label{tab:judgeresults}
\end{table}
\vskip -0.3in
\end{center}

We calculate the Agreement between the LLM evaluator and human evaluator for Vicuna-13b and GPT-3.5-Turbo separately. In the formula below, the numerator represents the number of instances where the LLM evaluator picks model A's response over model B, while the denominator represents the total number of instances where humans labeled model A as the better response.
$$
\text{Agreement\textsubscript{A}} = \frac{ \text{\texttt{Count}(LLM Prefers A)}}{\text{\texttt{Count}(Human Prefers A)}}
$$

The Bias Score, as given below, provides insight into potential bias in the LLM evaluator. If the difference is positive with a high magnitude, it indicates a bias toward Vicuna-13b, as the evaluator aligns more closely with human preferences for Vicuna-13b. Conversely, if the difference is negative with a high magnitude, it suggests a bias toward GPT-3.5-Turbo. A small magnitude (close to zero) implies that the LLM evaluator is relatively unbiased, showing similar levels of agreement with human preferences for both models.
$$
\text{Bias Score} = \text{A\textsubscript{Vicuna-13b}} - \text{A\textsubscript{GPT-3.5-Turbo}}
$$

The Bias Score highlights potential biases in the LLM evaluator's alignment with human preferences for Vicuna-13b and GPT-3.5-Turbo, as shown in Table \ref{tab:fair-eval_comparison}. Bias Score for Llama-3.1-8b (+0.51) and Llama-3.1-70b (+0.37), indicates a strong bias toward Vicuna-13b, where the evaluator more frequently favors Vicuna-13b over GPT-3.5-Turbo. Conversely, for Gemma-2-9b (-0.05) and Gemma-2-27b (+0.02) the small magnitude of Bias Score suggests that Gemma models are impartial. For GPT-4o-mini (+0.18) and GPT-4o (+0.33), the Bias Score indicates a moderate bias toward Vicuna-13b, as the evaluator shows a noticeable but less pronounced preference for Vicuna-13b's responses compared to GPT-3.5-Turbo. Vicuna-13b is fine-tuned on the ShareGPT dataset, which contains conversations from GPT-4 and GPT-3.5. This fine-tuning likely aligns Vicuna-13b's responses with GPT models, explaining the evaluator's bias toward it.

\begin{center}
\begin{table}[H]
\centering
\renewcommand{\arraystretch}{1}
\small
\begin{tabular}{l@{\hskip 0pt}c@{\hskip 6pt}c@{\hskip 4pt}c}
\toprule
\textbf{Model} & \multicolumn{2}{c}{\textbf{Agreement}} & \textbf{Bias Score} \\
\cmidrule(lr){2-3}
& \textbf{Vicuna-13b} & \textbf{GPT-3.5-Turbo} & \\
\midrule
Gemma-2-9b & 0.68 & 0.73 & -0.05 \\
Llama-3.1-8b & 0.92 & 0.41 & +0.51 \\
Gemma-2-27b & 0.80 & 0.78 & +0.02 \\
Llama-3.1-70b & 0.88 & 0.51 & +0.37 \\
GPT-4o-mini & 0.84 & 0.66 & +0.18 \\
GPT-4o & 0.92 & 0.59 & +0.33 \\
\bottomrule
\end{tabular}
\caption{\small{Agreement between the LLM evaluator and human evaluator over Vicuna-13b and GPT-3.5 separately.}}
\label{tab:fair-eval_comparison}
\end{table}
\vskip -0.3in
\end{center}

\begin{table*}[htbp]
\centering
\renewcommand{\arraystretch}{1}
\small 
\begin{tabular}{lcccccccccc}
\toprule
& \textbf{Fair-Eval} & \textbf{PandaLM-Eval} &\textbf{MT-Bench} \\
\textbf{Jury}  \\
\midrule
Gemma-2-9b, Gemma-2-27b, Llama-3.1-8b, GPT-4o-mini &  \textbf{0.428} & 0.604 & 0.395 \\
Gemma-2-9b, Gemma-2-27b, GPT-4o-mini, GPT-4o  & \underline{0.415} & 0.639 & 0.418  \\
Gemma-2-27b, Llama-3.1-70b, GPT-4o-mini, GPT-4o & 0.412 & 0.637 & 0.410 \\
Gemma-2-27b, GPT-4o-mini, GPT-4o & 0.396 & \textbf{0.673} & 0.400 \\
Llama-3.1-70b, GPT-4o-mini, GPT-4o & 0.365 & \underline{0.663} & 0.410 \\
Gemma-2-9b, GPT-4o-mini, GPT-4o & 0.375 & 0.662 & 0.416 \\
Llama-3.1-70b, GPT-4o & 0.273 & 0.636 & \textbf{0.429} \\
GPT-4o-mini, GPT-4o & 0.315 & 0.660 & \underline{0.426} \\
Gemma-2-9b, GPT-4o & 0.290 & 0.609 & 0.422 \\
\bottomrule
\end{tabular}
\caption{\small{Performance comparison of LLMs-as-a-Jury on the three datasets. For each dataset, we pick the top 3 juries. The bold score is for the best jury for the specific dataset and the underlined one is the second best.}}
\label{tab:juryresults}
\end{table*}

\paragraph{LLMs-as-a-Jury.} In evaluating of LLMs-as-a-Jury, we analyze the top three juries from each dataset as shown in Table \ref{tab:juryresults}. Notably, the scores exhibit considerable variation across the different datasets. On the Fair-Eval and MT-Bench datasets, the jury approach outperformed the judge approach, indicating a potential advantage in using multiple models for evaluation. For instance, on Fair-Eval, the highest-performing jury achieves a Cohen's Kappa of 0.428 while the judge achieves Kappa of 0.389, suggesting a relatively strong agreement with human judgments compared to individual judges. This configuration, however, shows a drop in performance on other datasets with a kappa of 0.604 on PandaLM-Eval and 0.395 on MT-Bench, underscoring the challenge of generalizing a single jury setup across varied datasets. However, the judge approach outperforms the jury on the PandaLM-Eval dataset, where the best judge attained a kappa of 0.688, surpassing the top jury's kappa of 0.673. The best jury on MT-Bench, with a kappa of 0.429, also demonstrates variability in its performance across datasets as well, with a kappa of 0.636 on PandaLM-Eval and only 0.273 on Fair-Eval.

The jury approach, by incorporating diverse models, mitigates the biases that occur in LLM-as-a-Judge approach when bench-marking on the Fair-Eval dataset. However while the jury approach can offer robustness through diversity, in evaluation task, it does not universally outperform single judges. The decision to employ a jury versus a judge should consider whether the candidate responses being evaluated include output from the judge itself, which can introduce bias in the results. Additionally, scalability should be taken into account, as the jury approach might require more computational resources. Another critical consideration is the variability in performance across different datasets, which poses a challenge for generalization.

\paragraph{LLM Debate.} The LLM Debate approach, as summarized in Table \ref{tab:debateresults}, showcases varying degrees of effectiveness across three different datasets: Fair-Eval, PandaLM-Eval, and MT-Bench. 

\begin{center}
\begin{table}[H]
\centering
\renewcommand{\arraystretch}{1}
\small
\begin{tabular}{lcccccccccc}
\toprule
& \textbf{Fair-Eval} & \textbf{PandaLM} &\textbf{MT-Bench} \\
\textbf{Debater}  \\
\midrule
Gemma-2-9b & 0.323 & 0.520 & 0.326 \\
Llama-3.1-8b & 0.080 & 0.440 & 0.309 \\
Gemma-2-27b & 0.336 & 0.605 & 0.363 \\
Llama-3.1-70b & 0.292 & 0.547 & 0.381 \\
GPT-4o-mini & 0.360 & 0.625 & 0.376 \\
GPT-4o & \textbf{0.404} & \textbf{0.654} & \textbf{0.402} \\
\bottomrule
\end{tabular}
\caption{\small{Performance comparison of LLM Debate on the three datasets.}}
\label{tab:debateresults}
\end{table}
\vskip -0.3in
\end{center}

GPT-4o performs the best across all datasets, with Cohen's Kappa scores of 0.404, 0.654, and 0.402 respectively. LLM Debate outperforms LLM-as-a-Judge on Fair-Eval only and does not surpass the LLMs-as-a-Jury approach on any dataset. On Fair-Eval using the Debate framework increases the Kappa score of GPT-4o from 0.327 to 0.404 and of GPT-4o-mini from 0.333 to 0.360. It shows that the debate approach decreases the bias of GPT-4o and GPT-4o-mini towards the responses of it's family. 

\begin{figure}[h!]
    \begin{center}
    \centerline{\includegraphics[width=1\columnwidth]{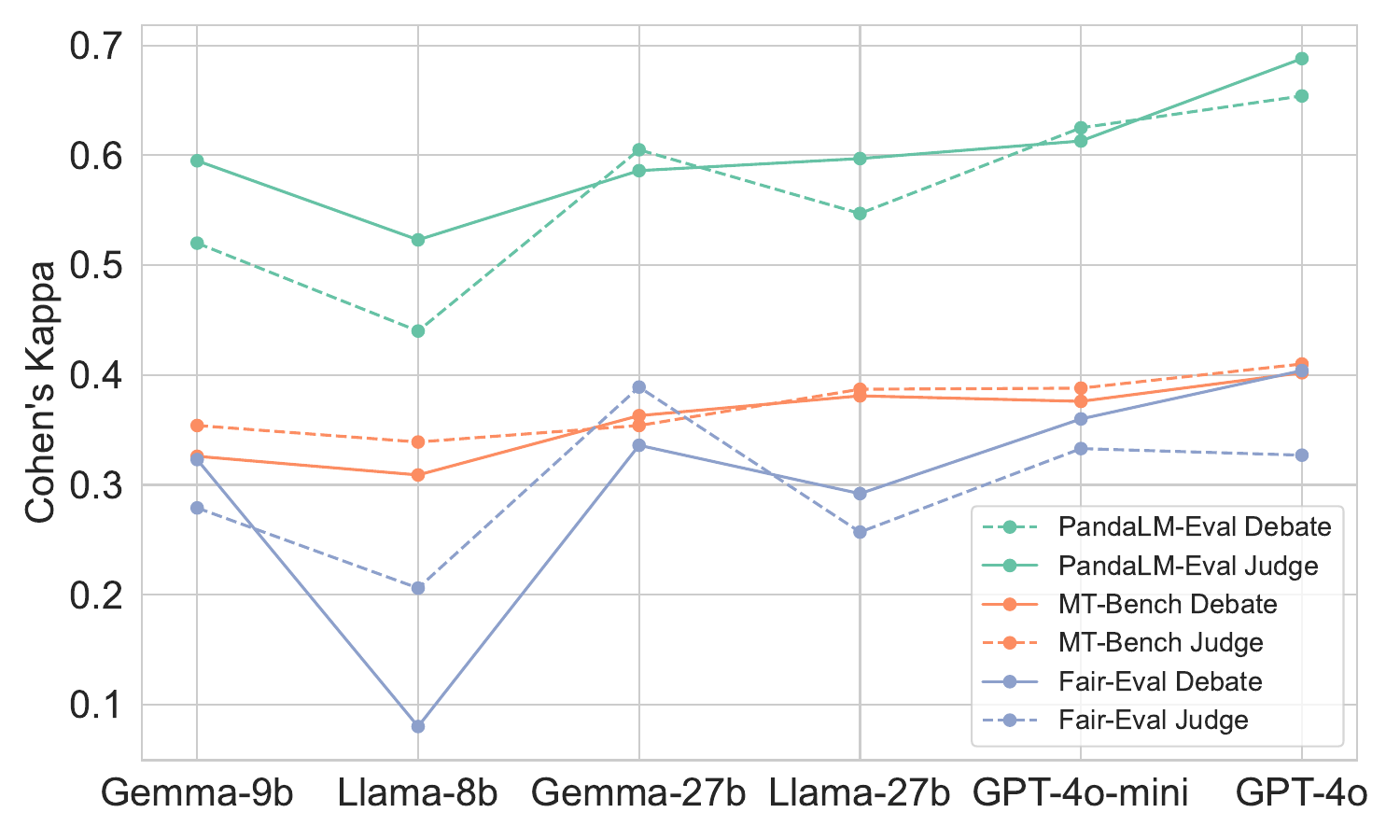}}
    \caption{\small{Comparison of LLM Debate and LLM-as-a-Judge across the three datasets and different models.}}
    \label{fig:debate_judge_comparison}
    \end{center}
    \vskip -0.3in
\end{figure}

There is a significant variance in the performance of LLM Debate across the models and the datasets. For instance, as seen in Table \ref{tab:debateresults} Gemma-2-27b in debate architecture outperforms Gemma-as-a-Judge on PandaLM-Eval and MT-Bench but on Fair-Eval judge performers better. Gemma-2-9b in debate architecture has a Kappa score of 0.323 on Fair-Eval, outperforming 0.279 of Gemma-as-a-Judge. However on PandaLM-Eval and MT-Bench Gemma-2-9b in debate framework achieves a Kappa score of 0.520 and 0.326, repectively. Both scores lower as compared to Gemma-as-a-Judge scores of 0.595 and 0.354. In case of Llama, Llama-3.1-8b in judge configuration outperforms itself in debate configuration. Llama-3.1-70b in debate framework only outperforms Llama-as-a-judge on Fair-Eval. Figure \ref{fig:debate_judge_comparison} shows a comparison of Cohen's Kappa of LLM Debate and LLM-as-a-Judge across the three datasets and all the models.

\paragraph{Evaluation Framework for PO Dataset.} Based on the comparative evaluation scores across the three datasets and the advantages and disadvantages associated with each multi-model framework, we have chosen to use the LLM-as-a-Judge approach with GPT-4o as our primary evaluator for generating the PO dataset. This decision is driven by multiple factors:
\begin{enumerate}
    \item In our context, the task involves generating a PO dataset using Llama-3.1-8b and Gemma-2-9b. Therefore there will be no bias in the evaluation when using GPT-4o as the judge.
    \item The performance of GPT-4o-as-a-Judge has been consistently high across various evaluations, indicating its reliability as a judge. While the LLMs-as-a-Jury and LLM Debate approaches have a high variance in Cohen's Kappa score across different datasets.
    \item The computational resources required for managing the LLM Debate and LLM Jury frameworks are considerably higher than those needed for a single-judge setup. The LLM-as-a-Judge method is simpler to implement and scale.
\end{enumerate}

\subsection{LLM-as-Generator}
We compare the performance of multi-model Feedback Loop with the baseline single-models (GPT-4o, Llama-3.1-8b, Gemma-2-9b) using win rate as shown in Table \ref{tab:winrates}.

\begin{center}
\begin{table}[htbp]
\centering
\renewcommand{\arraystretch}{1}
\small
\begin{tabular}{llccccccccc}
\toprule
& & \multicolumn{3}{c}{\textbf{Win Rate (\%) Against}} \\
\cmidrule(lr){3-5}
\textbf{Generator} & \textbf{Reviewer} & \textbf{GPT} & \textbf{Llama} & \textbf{Gemma} \\
\midrule
Gemma & - & 38.6 & 66.6 & - \\
Llama & - & 39.2 & - & 33.4 \\
\midrule
Gemma & Gemma & 41.4 & 64.8 & 52.6 \\
 & Llama & 41.2 & 61.8 & 47.8 \\
 & Both & 42.0 & 67.6 & 52.4 \\
\midrule
Llama & Gemma & \textbf{49.0} & \textbf{71.8} & \textbf{73.8} \\
 & Llama & 47.8 & 65.8 & 65.6 \\
 & Both & 48.6 & 68.2 & 69.4 \\
\bottomrule
\end{tabular}
\caption{\small{Win Rate of multi-model and single-model against GPT-4o, Llama-3.1-8b and Gemma-2-9b}}
\label{tab:winrates}
\end{table}
\vskip -0.3in
\end{center}

We utilize GPT-4o-as-a-judge in this evaluation process. For the baseline we find the win rate of Gemma and Llama against GPT-4o and each other. Both smaller models have similar win rate of 38.6\% and 39.2\% against GPT, while Gemma has a win rate of 66.6\% against Llama.

In the multi-model setting, all variations outperform the single-models against GPT-4o, with the highest win rate of 49.0\% for Llama as a generator and Gemma as a reviewer. This configuration performs the best against Llama and Gemma too, with 71.8\% and 73.8\% win rate respectively. We observe that using Llama as the generator improves the performance as compared to using Gemma as the generator because this configuration leads to a better win rate against all three baselines. 

Llama's strengths in generating responses may be enhanced by Gemma's ability to fine-tune and correct the errors, leading to more polished outputs. The results underscore the importance of assigning appropriate roles based on the specific strengths of each model. Llama, when set as the generator, appears to leverage its capabilities more effectively than Gemma in this role. The use of diverse models in the feedback loop likely helps mitigate biases that any single model might introduce. This diversity ensures a broader range of perspectives while answer a question. In conclusion, the demonstrated efficacy of the multi-model Feedback Loop, especially with Llama as the generator and Gemma as the reviewer, validates the concept of collaborative AI systems.

\subsection{Preference Optimization Dataset}
\begin{figure}[h!]
    \begin{center}
    \centerline{\includegraphics[width=0.8\columnwidth]{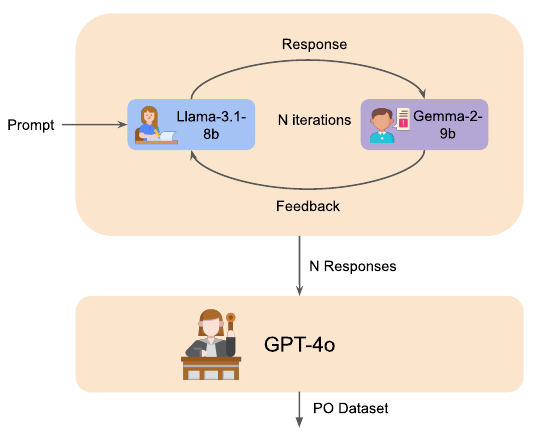}}
    \caption{\small{Multi-model framework for PO dataset generation.}}
    \label{fig:final}
    \end{center}
    \vskip -0.3in
\end{figure}

We use GPT-4o-as-a-Judge in the evaluation module because of its consistency and reliability as a judge across multiple datasets. In the generation module, we use LLM Feedback Loop with Llama-3.1-8b as the generator and Gemma-2-9b as the reviewer because of it's highest win-rate against other configurations. The framework is shown in Figure \ref{fig:final}. For the dataset generation, we use $N=3$ feedback iterations. For each prompt, we generate three responses using the generation module. These responses are then evaluated by GPT-4o in the evaluation module. The response judged to be the best by GPT-4o is labeled as accepted, while the other two responses are labeled as rejected to form the DPO and KTO datasets.

\section{Conclusion}
This paper presents PO datasets generated using multi-model frameworks, and evaluates these frameworks by highlighting the advantages, drawbacks, and challenges of each approach. In the response evaluation module, our comparative analysis of LLM-as-a-Judge, LLMs-as-a-Jury, and LLM Debate shows the suitability of each setup depending on the context of use. For the response generation module, we evaluate the LLM Feedback Loop using Llama-3.1-8b and Gemma-2-9b in various configurations. LLM-as-a-Judge proved to be highly effective when candidate responses don’t have a response from the Judge LLM. Whereas LLMs-as-a-Jury and LLM Debate demonstrated robustness, particularly useful in reducing evaluator bias. However, Cohen’s Kappa for both of these approaches has a high variance making them less suitable for novel applications.

Our experiments with LLM Feedback Loop using Llama-3.1-8b and Gemma-2-9b configurations show the potential of multi-model frameworks in refined content generation. Configurations where Llama-3.1-8b served as the generator and Gemma-2-9b as the reviewer consistently delivered better results, demonstrating the benefits of leveraging complementary strengths of different models to refine output quality. These findings indicate the effectiveness of multi-model frameworks for varied AI applications, showing promise for moving towards systems requiring minimal human intervention - however, this method is computationally expensive in comparison.

We also generate multiple DPO and KPO datasets using LLM Feedback Loop with Llama-3.1-8b as the generator and Gemma-2-9b as the evaluator and GPT-4o-as-a-Judge. The aim of these datasets is to improve single-model capabilities for better response generation and multi-model capabilities including better communication and improved feedback. 

\section{Future Work}
In terms of future work, there are three avenues of investigation: 
(1) Performance comparison of models fine-tuned on our PO dataset versus widely-used LLMs to investigate the impact of our generated datasets through a series of experiments. (2) Using larger models such as Llama-3.1-70b and Gemma-2-27b for dataset generation as this may provide more diverse and higher-quality training data, potentially leading to further advancements in model performance and generalizability. (3) Experimenting with the number of iterations used in the Feedback Loop framework and including other LLM families in the dataset generation process.

\section*{Limitations}
While our study demonstrates the potential of multi-model workflows in automating the generation of PO datasets, several limitations should be acknowledged. Firstly, the use of multi-model frameworks significantly increases computational complexity and resource consumption compared to single-model models. The iterative processes in both the response generation and evaluation modules require more computational power and time, which may not be feasible for practitioners with limited resources. Additionally, GPT-4o is a proprietary model, which may not be accessible to all researchers, potentially hindering reproducibility and wider adoption of our methods.

\section*{Ethical Considerations}
The automation of response evaluation and generation in PO dataset creation raises several ethical considerations that warrant careful attention. Relying on LLMs to simulate human judgments may perpetuate existing biases present in the training data of these models. If not properly addressed, this could result in PO datasets that reinforce stereotypes or unfairly represent certain groups, leading to biased behaviors in models fine-tuned on these datasets. The potential displacement of human annotators poses an ethical dilemma. While automation can increase efficiency and scalability, it may reduce opportunities for human involvement in the annotation process, affecting those who rely on such tasks for employment. Balancing automation with human oversight is essential to maintain ethical standards and ensure diverse perspectives are included.

In conclusion, while our approach offers advancements in automating PO dataset generation, it is imperative to remain vigilant about these ethical concerns. Implementing strategies to mitigate biases, maintaining transparency, involving human oversight, and adhering to ethical guidelines are essential steps in responsible AI development.


\bibliography{custom}

\appendix
\section{System Prompts}
\label{sec:system-prompts}
Table \ref{tab:prompt-llm-judge} contains the three categories of system prompts tested for LLM-as-a-Judge approach. The winning prompt with \textit{Combined Scoring} was used for LLMs-as-a-Jury. These prompts are modified versions of those used by \citep{zheng2023judgingllmasajudgemtbenchchatbot}. Table \ref{tab:prompt-llm-debate} present the system prompt and user message structure for LLM Debate and \ref{tab:prompt-roles} shows the prompt for each role in the debate. This is based on the system prompt and the input structure used by \citep{chan2023chatevalbetterllmbasedevaluators}. Table \ref{tab:generator_prompt} shows the user message structure for the generator LLM and Table \ref{tab:reviewer_prompt} shows the system prompt and user message for reviewer LLM in LLM Feedback Loop.

\section{Code and Datasets}
The evaluation code, all the evaluation outputs and the generated datasets are publicly available on GitHub\footnote{\url{https://github.com/sameearif/Fellowship-of-The-LLMs}}. For evaluation of LLMs-as-Evaluators we used Alpaca-Eval\footnote{\url{https://huggingface.co/datasets/tatsu-lab/alpaca_eval}}, Fair-Eval\footnote{\url{https://github.com/i-Eval/FairEval}}, PandaLM-Eval\footnote{\url{https://github.com/WeOpenML/PandaLM}} and MT-Bench\footnote{\url{https://huggingface.co/datasets/lmsys/mt_bench_human_judgments}}. For evaluation of LLMs-as-Generators and single-model improvement dataset generation we use the prompts from Argilla Capybara DPO dataset\footnote{\url{https://huggingface.co/datasets/argilla/distilabel-capybara-dpo-7k-binarized}}. For multi-model improvement dataset generation we use prompts from No-Robots\footnote{\url{https://huggingface.co/datasets/HuggingFaceH4/no_robots}} dataset. Alpaca-Eval and PandaLM-Eval are under Apache 2.0 license, Fair-Eval dataset is under CC BY 4.0 license, Argilla Capybara DPO is also under Apache 2.0 license. All datasets used in this paper comply with their respective license.

\section{Computing Infrastructure}
We use the API for GPT-4o and GPT-4o-mini from OpenAI\footnote{\url{https://platform.openai.com/docs/overview}}. For Gemma and Llama models API from TogetherAI\footnote{\url{https://docs.together.ai/docs/introduction}} was used. We use Python3 libraries for both APIs and the temperature for the models was set to 0 for reproduciblity. For each evaluation, one run of the code was done. OpenAI GPT-4o has a proprietary license. Llama-3.1 is under Llama-3.1 license and Gemma-2 is under Gemma license. All models used in this paper comply with their respective license.

\renewcommand{\arraystretch}{1.5}

\onecolumn
\begin{longtable}{|p{1.2in}|p{4.7in}|}
\caption{\small{The three types of system prompts for LLM-as-a-Judge and LLMs-as-a-Jury.}}
\label{tab:prompt-llm-judge}\\
\hline
\textbf{Prompt Type} & \textbf{Prompt} \\
\hline
\endfirsthead

\caption[]{\small{The three types of system prompts for LLM-as-a-Judge and LLMs-as-a-Jury (continued).}} \\
\hline
\textbf{Prompt Type} & \textbf{Prompt} \\
\hline
\endhead

\hline
\endfoot

\hline
\endlastfoot

Direct Comparison & Please act as an impartial judge and evaluate the quality of the responses provided by two AI assistants to the user question displayed below. You should choose the assistant that follows the user's instructions and answers the user's questions better. Your evaluation should consider factors such as the helpfulness, relevance, accuracy, depth, creativity, and level of detail of their responses. Begin your evaluation by comparing the two responses and provide a short explanation. Avoid any position biases and ensure that the order in which the responses were presented does not influence your decision. Do not allow the length of the responses to influence your evaluation. Answer options: \newline
A: If response by assistant A is better \newline
B: If response by assistant B is better \newline
C: If it is a tie \newline \newline
Use the following format to respond: \newline
\#\#\# Evaluation Evidence:  \newline
[Add your explanation here] \newline \newline
\#\#\# Answer: \newline
A or B or C \\
\hline
Independent Scoring & Please act as an impartial judge and evaluate the quality of the response provided by an AI assistant to the user question displayed below. Assign an overall score out of 10, where a higher score indicates better overall performance. Your evaluation should consider factors such as the helpfulness, relevance, accuracy, depth, creativity, and level of detail of their response. Begin your evaluation by comparing the two responses and provide a short explanation. Do not allow the length of the response to influence your evaluation. \newline \newline
Use the following format to respond: \newline 
\#\#\# Evaluation Evidence: \newline
[Add your explanation here] \newline \newline
\#\#\# Overall Score: \newline
X/10 \\
\hline
Combined Scoring & Please act as an impartial judge and evaluate the quality of the responses provided by two AI assistants to the user question displayed below. You should choose the assistant that follows the user's instructions and answers the user's questions better. Each response receives an overall score out of 10, where a higher score indicates better overall performance. Your evaluation should consider factors such as the helpfulness, relevance, accuracy, depth, creativity, and level of detail of their responses. Begin your evaluation by comparing the two responses and provide a short explanation. Avoid any position biases and ensure that the order in which the responses were presented does not influence your decision. Do not allow the length of the responses to influence your evaluation. \newline \newline
Use the following format to respond: \newline
\#\#\# Evaluation Evidence: \newline
[Add your explanation here] \newline \newline
\#\#\# Score Assistant A: \newline
X/10 \newline \newline
\#\#\# Score Assistant B: \newline
Y/10 \\
\end{longtable}

\begin{longtable}{|p{1.2in}|p{4.7in}|}
\caption{\small{The system prompt and the user message structure for LLM Debate.}}\\
\hline
\textbf{Message Type} & \textbf{Prompt} \\
\hline
\endfirsthead

\caption[]{\small{The system prompt and the user message structure for LLM Debate (continued).}} 
\label{tab:prompt-llm-debate}\\
\hline
\textbf{Message Type} & \textbf{Prompt} \\
\hline
\endhead

\hline
\endfoot

\hline
\endlastfoot

System Prompt & We would like to request your feedback on the performance of two AI assistants in response to the user question. There are a few other referees assigned the same task; it's your responsibility to discuss with them and think critically before you make your final judgement. \newline
Each response receives an overall score on a scale of 1 to 10, where a higher score indicates better overall performance. You should choose the assistant that follows the user's instructions and answers the user's question better. You don't necessarily have to agree with others. \newline
Your evaluation should consider factors such as the helpfulness, relevance, accuracy, depth, creativity, and level of detail of their responses. Avoid any position biases and ensure that the order in which the responses were presented does not influence your decision. Do not allow the length of the responses to influence your evaluation. \\
\hline
User Message & \texttt{<|}Start of User Question\texttt{|>}\newline \{User Question\}\newline \texttt{<|}End of User Question\texttt{|>}\newline \newline \texttt{<|}The Start of Assistant 1's Answer\texttt{|>}\newline \{Assistant 1\}\newline \texttt{<|}The End of Assistant 1's Answer\texttt{|>}\newline \newline \texttt{<|}The Start of Assistant 2's Answer\texttt{|>}\newline \{Assistant 2\}\newline \texttt{<|}The End of Assistant 2's Answer\texttt{|>}\newline \newline Here is your discussion history:\newline \{Chat History\}\newline\newline \{Role\} \\
\end{longtable}

\begin{longtable}{|p{1.2in}|p{4.7in}|}
\caption{\small{The prompt for each role used in LLM Debate.}}
\label{tab:prompt-roles}\\
\hline
\textbf{Role} & \textbf{Prompt} \\
\hline
\endfirsthead

\caption[]{\small{The prompt for each role used in LLM Debate (continued).}} \\
\hline
\textbf{Role} & \textbf{Prompt} \\
\hline
\endhead

\hline
\endfoot

\hline
\endlastfoot

General Public & You are now General Public, one of the referees in this task. You are interested in the story and looking for updates on the investigation. Please think critically by yourself and note that it's your responsibility to choose which of the responses is better first. \newline \newline Now it's your turn to speak, General Public. Please make your talk short and clear. \newline **General Public**: \\
\hline
Psychologist & You are now Psychologist, one of the referees in this task. You will study human behavior and mental processes in order to understand and explain human behavior. Please help others determine which response is the better one. \newline \newline Now it's your turn to speak, Psychologist. Please make your talk short and clear. \newline **Psychologist**: \\
\hline
Critic & You are now Critic, one of the referees in this task. You will check for fluent writing, clear sentences, and good wording in summary writing. Your job is to question others' judgment to make sure their judgment is well-considered and offer an alternative solution if two responses are at the same level. \newline \newline Now it's your turn to speak, Critic. Please make your talk short and clear. \newline **Critic**: \\
\end{longtable}

\begin{longtable}{|p{1.2in}|p{4.7in}|}
\caption{\small{The user message structure for the generator in LLM Feedback.}} \label{tab:generator_prompt} \\
\hline
\textbf{Message Type} & \textbf{Prompt} \\
\hline
\endfirsthead

\caption[]{\small{The user message structure for the generator in LLM Feedback (continued).}} \\
\hline
\textbf{Message Type} & \textbf{Prompt} \\
\hline
\endhead

\hline
\endfoot

\hline
\endlastfoot

User Message (Single Feedback) & Update your response based on the feedback: \newline [Start of Feedback] \newline \{Feedback\} \newline [End of Feedback] \newline \newline Do not engage in formalities such as 'Thank you for your feedback' or 'Here is an updated version...' etc., just update the response. \\
\hline
User Message (Double Feedback) & Update your response based on the feedback by the two assistants: \newline [Start of Assistant 1's Feedback] \newline \{Assistant 1's Feedback\} \newline [End of Assistant 1's Feedback] \newline \newline [Start of Assistant 2's Feedback] \newline \{Assistant 2's Feedback\} \newline [End of Assistant 2's Feedback] \newline \newline Do not engage in formalities such as 'Thank you for your feedback' or 'Here is an updated version...' etc., just update the response. \\
\end{longtable}

\begin{longtable}{|p{1.2in}|p{4.7in}|}
\caption{\small{The prompt and user message structure for the reviewer in LLM Feedback.}} \\
\hline
\textbf{Message Type} & \textbf{Prompt} \\
\hline
\endfirsthead

\caption[]{\small{The prompt and user message structure for the reviewer in LLM Feedback (continued).}} 
\label{tab:reviewer_prompt}\\
\hline
\textbf{Message Type} & \textbf{Prompt} \\
\hline
\endhead

\hline
\endfoot

\hline
\endlastfoot

System Prompt & Please give constructive feedback on how to improve the response provided by an AI assistant to the user question. \newline
Your evaluation should consider factors such as the instruction following (the response should align with the user instructions), helpfulness, relevance, accuracy, and creativity of the response. \newline
Assign an overall score out of 10, up to one decimal place, where a higher score indicates better overall performance. \newline \newline
Use the following format to respond: \newline
\#\#\# Evaluation: \newline
[Add your evaluation here] \newline \newline
\#\#\# Overall Score: \newline
X/10 \newline \newline
\#\#\# Feedback: \newline
[Add your feedback here] \\
\hline
User Message & [Start of User Question] \newline \{User Question\} \newline [End of User Question] \newline \newline [Start of Assistant's Response] \newline \{Assistant's Response\} \newline [End of Assistant's Response] \\
\end{longtable}

\end{document}